\documentclass[letterpaper, 10 pt, conference]{ieeeconf}  

\IEEEoverridecommandlockouts                              

\usepackage{graphics} 
\usepackage{epsfig} 
\usepackage{mathptmx} 
\usepackage{times} 
\usepackage{amsmath} 
\usepackage{amssymb}  
\usepackage{multirow}
\usepackage{subcaption}
\usepackage{color}
\usepackage{xspace}
\usepackage{xfrac}
\usepackage{hyperref}
\usepackage[keeplastbox]{flushend} 
\makeatletter
\DeclareRobustCommand\onedot{\futurelet\@let@token\@onedot}
\def\@onedot{\ifx\@let@token.\else.\null\fi\xspace}

\def\eg{\emph{e.g}\onedot}

\def\wrt{w.r.t\onedot} 
\def\etal{\emph{et al}\onedot}
\makeatother

\newcommand{\bs}[1]{\boldsymbol{\mathbf{#1}}}
\title{\LARGE \bf
Using Simulation and Domain Adaptation to Improve\\
Efficiency of Deep Robotic Grasping
}

\author{Konstantinos Bousmalis$^{*,1}$, Alex Irpan$^{*,1}$, Paul Wohlhart$^{*,2}$, Yunfei Bai$^{2}$, Matthew Kelcey$^{1}$, Mrinal Kalakrishnan$^{2}$,\\ Laura Downs$^{1}$, Julian Ibarz$^{1}$, Peter Pastor$^{2}$, Kurt Konolige$^{2}$,  Sergey Levine$^{1}$, Vincent Vanhoucke$^{1}$
\thanks{*Authors contributed equally, $^{1}$Google Brain, $^{2}$X }
}

\begin{document}

\maketitle
\thispagestyle{empty}
\pagestyle{empty}

\begin{abstract}
Instrumenting and collecting annotated visual grasping datasets to train modern machine learning algorithms can be extremely time-consuming and expensive. An appealing alternative is to use off-the-shelf simulators to render synthetic data for which ground-truth annotations are generated automatically. 
Unfortunately, models trained purely on simulated data often fail to generalize to the real world.
We study how randomized simulated environments and domain adaptation methods can be extended to train a grasping system to grasp novel objects from raw monocular RGB images.
We extensively evaluate our approaches with a total of more than 25,000 physical test grasps, studying a range of simulation conditions and domain adaptation methods, including a novel extension of pixel-level domain adaptation that we term the GraspGAN.
We show that, by using synthetic data and domain adaptation, we are able to reduce the number of real-world samples needed to achieve a given level of performance by up to 50 times, using only randomly generated simulated objects. We also show that by using only unlabeled real-world data and our GraspGAN methodology, we obtain real-world grasping performance without any real-world labels that is similar to that achieved with 939,777 labeled real-world samples.
\end{abstract}

\section{Introduction}

Grasping is one of the most fundamental robotic manipulation problems. For virtually any prehensile manipulation behavior, the first step is to grasp the object(s) in question. Grasping has therefore emerged as one of the central areas of study in robotics, with a range of methods and techniques from the earliest years of robotics research to the present day. A central challenge in robotic manipulation is generalization: can a grasping system successfully pick up diverse new objects that were not seen during the design or training of the system? Analytic or model-based grasping methods~\cite{handbook_of_robotics} can achieve excellent generalization to  situations that satisfy their assumptions. However, the complexity and unpredictability of unstructured real-world scenes has a tendency to confound these assumptions, and learning-based methods have emerged as a powerful complement~\cite{bohg2014data,kappler,viereck2017learning,gupta,levine2016learning}.

\begin{figure}[t]
\centering
\includegraphics[width=\linewidth]{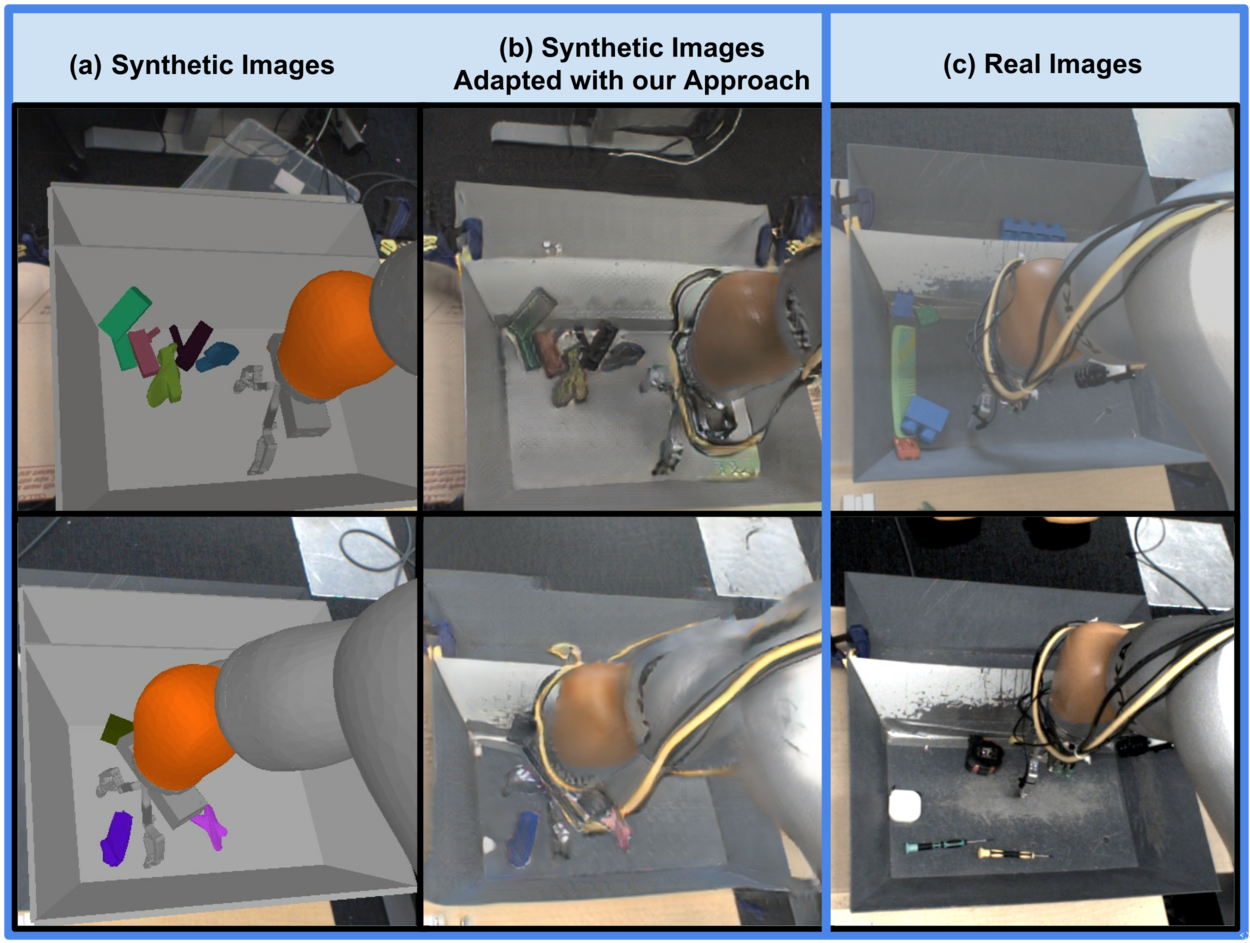}
\caption{\textbf{Bridging the reality gap:} our proposed pixel-level domain adaptation model takes as input \textsl{(a)} synthetic images produced by our simulator and produces \textsl{(b)} adapted images that look similar to \textsl{(c)} real-world ones produced by the camera over the physical robot's shoulder. We then train a deep vision-based grasping network with adapted and real images, which we further refine with feature-level adaptation.}
\label{fig:intro_fig}
\vspace{-0.28in}
\end{figure}

Learning a robotic grasping system has the benefit of generalization to objects with real-world statistics, and can benefit from the advances in computer vision and deep learning. Indeed, many of the grasping systems that have shown the best generalization in recent years incorporate convolutional neural networks into the grasp selection process~\cite{bohg2014data,gupta,viereck2017learning,mahler2017dex}. However, learning-based approaches also introduce a major challenge: the need for large labeled datasets. These labels might consist of human-provided grasp points~\cite{lenz2015}, or they might be collected autonomously~\cite{gupta,levine2016learning}. In both cases, there is considerable cost in both time and money, and recent studies suggest that the performance of grasping systems might be strongly influenced by the amount of data available~\cite{levine2016learning}.

A natural avenue to overcome these data requirements is to look back at the success of analytic, model-based grasping methods~\cite{handbook_of_robotics}, which incorporate our prior knowledge of physics and geometry. We can incorporate this prior knowledge into a learning-based grasping system in two ways. First, we could modify the design of the system to use a model-based grasping method, for example as a scoring function for learning-based grasping~\cite{mahler2017dex}. Second, we could use our prior knowledge to construct a simulator, and generate synthetic experience that can be used in much the same way as real experience. The second avenue, which we explore in this work, is particularly appealing because we can use essentially the same learning system.
However, incorporating simulated images presents challenges: simulated data differs in systematic ways from real-world data, and  simulation must have sufficiently general objects. Addressing these two challenges is the principal subject of our work.

Our work has three main contributions. \textit{(a) Substantial improvement in grasping performance from monocular RGB images by incorporating synthetic data:}
We propose approaches for incorporating synthetic data into end-to-end training of vision-based robotic grasping that we show achieves substantial improvement in performance, particularly in the lower-data and no-data regimes. 
\textit{(b) Detailed experimentation for simulation-to-real world transfer:} 
Our experiments involved $25,704$ real grasps of $36$ diverse test objects and consider a number of dimensions: the nature of the simulated objects, the kind of randomization used in simulation, and the domain adaptation technique used to adapt simulated images to the real world. 
\textit{(c) The first demonstration of effective simulation-to-real-world transfer for purely monocular vision-based grasping:}
To our knowledge, our work is the first to demonstrate successful simulation-to-real-world transfer for grasping, with generalization to previously unseen natural objects, using only monocular RGB images.

\section{Related Work}
\label{sect:related}
\textbf{Robotic grasping} is one of the most widely explored areas of manipulation. While a complete survey of grasping is outside the scope of this work, we refer the reader to standard surveys on the subject for a more complete treatment~\cite{bohg2014data}. 
Grasping methods can be broadly categorized into two groups: geometric methods and data-driven methods. Geometric methods employ analytic grasp metrics, such as force closure~\cite{force_closure_1995} or caging~\cite{caging_servey}. These methods often include appealing guarantees on performance, but typically at the expense of relatively restrictive assumptions. Practical applications of such approaches typically violate one or more of their assumptions. For this reason, data-driven grasping algorithms have risen in popularity in recent years. Instead of relying exclusively on an analytic understanding of the physics of an object, data-driven methods seek to directly predict either human-specified grasp positions~\cite{lenz2015} or empirically estimated grasp outcomes~\cite{gupta,levine2016learning}. A number of methods combine both ideas, for example using analytic metrics to label training data~\cite{kappler,mahler2017dex}.

\textbf{Simulation-to-real-world transfer in robotics} is an important goal, as simulation can be a source of practically infinite cheap data with flawless annotations. For this reason, a number of recent works have considered simulation-to-real world transfer in the context of robotic manipulation. Saxena \etal~\cite{saxena2008robotic} used rendered objects to learn a vision-based grasping model.
Gulatieri \etal and Viereck \etal ~\cite{viereck2017learning, gualtieri2016high} have considered simulation-to-real world transfer using depth images. Depth images can abstract away many of the challenging appearance properties of real-world objects. However, not all situations are suitable for depth cameras, and coupled with the low cost of simple RGB cameras, there is considerable value in studying grasping systems that solely use monocular RGB images. 

A number of recent works have also examined using randomized simulated environments~\cite{tobin2017domain,james2017transferring} for simulation-to-real world transfer for grasping and grasping-like manipulation tasks, extending on prior work on randomization for robotic mobility~\cite{sadeghi2016cad2rl}. These works apply randomization in the form of random textures, lighting, and camera position to their simulator. However, unlike our work, these prior methods considered grasping in relatively simple visual environments, consisting of cubes or other basic geometric shapes, and have not yet been demonstrated on grasping diverse, novel real-world objects of the kind considered in our evaluation.

\textbf{Domain adaptation} is a process that allows a machine learning model trained with samples from a source domain to generalize to a target domain. In our case the source domain is the simulation, whereas the target is the real world. 
There has recently been a significant amount of work on domain adaptation, particularly for computer vision~\cite{patel2015visual, csurka2017domain}.
Prior work can be grouped into two main types: feature-level and pixel-level adaptation.
\textit{Feature-level domain adaptation} focuses on learning domain-invariant features, either by learning a transformation of fixed, pre-computed features between source and target domains~\cite{sun2015return,gong2012geodesic,caseiro2015beyond,gopalan2011domain} or by learning a domain-invariant feature extractor, often represented by a convolutional neural network (CNN) ~\cite{ganin2016domain, long2015learning,  bousmalis2016domain}. Prior work has shown the latter is empirically preferable on a number of classification tasks~\cite{ganin2016domain, bousmalis2016domain}. Domain-invariance can be enforced by optimizing domain-level similarity metrics like maximum mean discrepancy~\cite{bousmalis2016domain}, or the response of an adversarially trained domain discriminator~\cite{ganin2016domain}. 
\textit{Pixel-level domain adaptation} focuses on re-stylizing images from the source domain to make them look like images from the target domain \cite{taigman2017unsupervised, bousmalis2017pixelda, shrivastava2017learning, zhu2017unpaired}. 
To our knowledge, all such methods are based on image-conditioned generative adversarial networks (GANs)~\cite{goodfellow2014generative}.
In this work, we compare a number of different domain adaptation regimes.
We also present a new method that combines both feature-level and pixel-level domain adaptation for simulation-to-real world transfer for vision-based grasping. 

\section{Background}
Our goal in this work is to show the effect of using simulation and domain adaptation in conjunction with a tested 
data-driven, monocular vision-based grasping approach. To this effect, we use such an approach, as recently proposed by Levine \etal~\cite{levine2016learning}. In this section we will concisely discuss this approach, and the two main domain adaptation techniques~\cite{ganin2016domain,bousmalis2017pixelda,shrivastava2017learning} our method is based on.
\label{sec:background}
\subsection{Deep Vision-Based Robotic Grasping}
\label{sec:hand-eye}
The grasping approach~\cite{levine2016learning} we use in this work consists of two components. The first is {\bf a grasp prediction convolutional neural network} (CNN)
$C({\bs x}_{i}, {\bs v}_i)$ that accepts a tuple of visual inputs ${\bs x}_i = \{{\bs x}_{i_0}, {\bs x}_{i_c}\}$ and a motion command ${\bs v}_i$, and outputs the predicted probability that executing ${\bs v}_i$ will result in a successful grasp. ${\bs x}_{i_0}$ is an image recorded before the robot becomes visible and starts the grasp attempt, and ${\bs x}_{i_c}$ is an image recorded at the current timestep. ${\bs v}_i$ is specified in the frame of the base of the robot and corresponds to a relative change of the end-effector's current position and rotation about the vertical axis. We consider only top-down pinch grasps, and the motion command has, thus, 5 dimensions: 3 for position, and 2 for a sine-cosine encoding of the rotation. The second component of the method is {\bf a simple, manually designed servoing function} that uses the grasp probabilities predicted by $C$ to choose the motor command ${\bs v}_i$ that will continuously control the robot. We can train the grasp prediction network $C$ using standard supervised learning objectives, and so it can be optimized independently from the servoing mechanism. In this work, we focus on extending the first component to include simulated data in the training set for the grasp prediction network $C$, leaving the other parts of the system unchanged.

The datasets for training the grasp prediction CNN $C$ are collections of visual episodes of robotic arms attempting to grasp various objects.
Each grasp attempt episode consists of $T$ time steps which result in $T$ distinct training samples. Each sample $i$ includes ${\bs x}_{i}, {\bs v}_i$, and the success label $y_i$ of the entire grasp sequence. 
The visual inputs are $640 \times 512$ images that are randomly cropped to a $472\times 472$ region during training to encourage translation invariance.

\begin{figure}[t]
\centering
\begin{subfigure}[b]{0.45\linewidth}
\includegraphics[width=\linewidth,trim={0 15cm 0 0},clip]{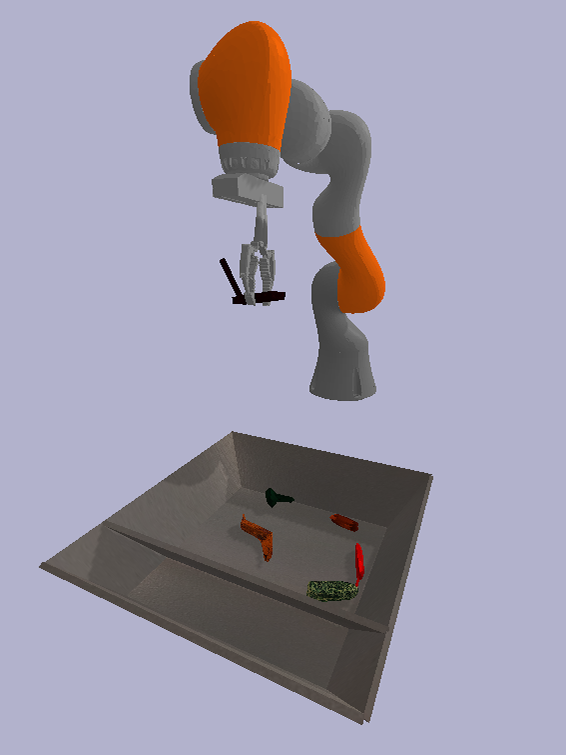}    
\caption{Simulated World}
\end{subfigure}
\begin{subfigure}[b]{0.45\linewidth}
\includegraphics[width=\linewidth,trim={0 0 0 4cm},clip]{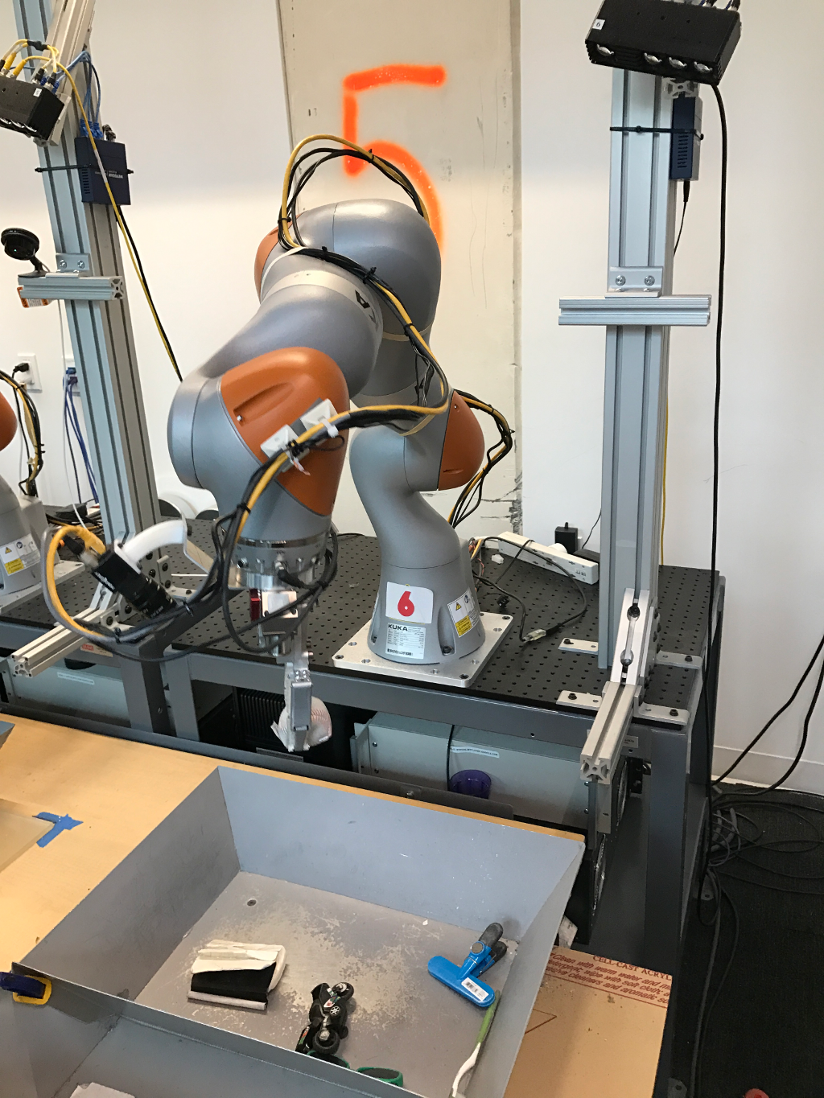}    
\caption{Real World}
\end{subfigure}\\
\begin{subfigure}[b]{0.45\linewidth}
\includegraphics[width=\textwidth]{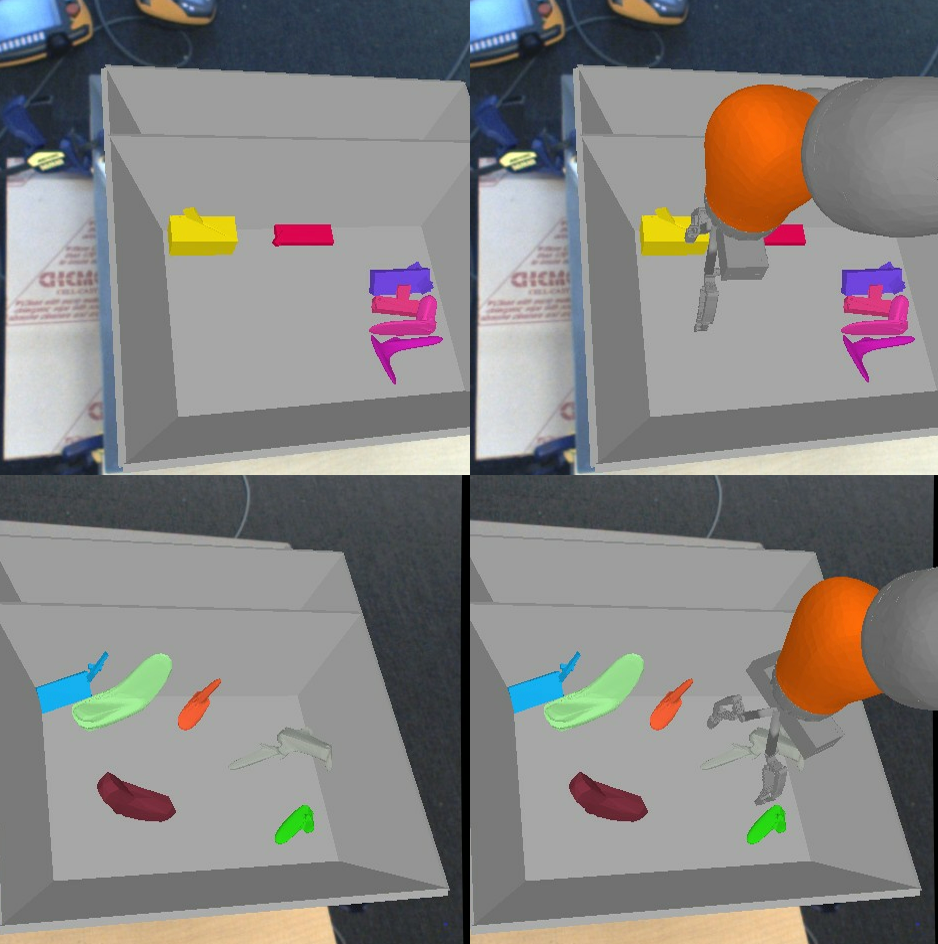}
\caption{Simulated Samples}
\end{subfigure}
\begin{subfigure}[b]{0.45\linewidth}
\includegraphics[width=\textwidth]{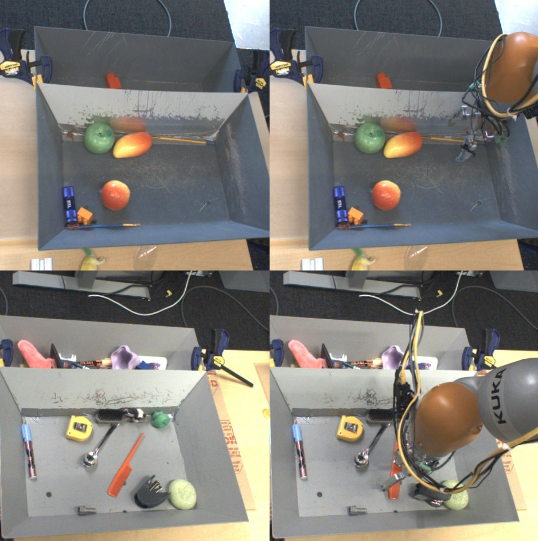}
\caption{Real Samples}
\end{subfigure}
\caption{{\bf Top Row:} The setup we used for collecting the \textsl{(a)} simulated and \textsl{(b)} real-world datasets. {\bf Bottom Row:} Images used during training of \textsl{(c)} simulated grasping experience with procedurally generated objects; and of \textsl{(d)} real-world experience with a varied collection of everyday physical objects. In both cases, we see the pairs of image inputs for our grasp success prediction model $C$: the images at $t=0$ and the images at the current timestamp. \vspace{-5mm}}
\label{fig:procedural_training}
\end{figure}

The central aim of our work is to compare different training regimes that combine both simulated and real-world data for training $C$. Although we do consider training entirely with simulated data, as we discuss in Section~\ref{sect:methods}, most of the training regimes we consider combine medium amounts of real-world data with large amounts of simulated data. To that end, we use the self-supervised real-world grasping dataset collected by Levine \etal~\cite{levine2016learning} using 6 physical Kuka IIWA arms. The goal of the robots was to grasp any object within a specified goal region. Grasping was performed using a compliant two-finger gripper picking objects out of a metal bin, with a monocular RGB camera mounted behind the arm. The full dataset includes about 1 million grasp attempts on approximately $1,100$ different objects, resulting in about 9.4 million real-world images.  
About half of the dataset was collected using random grasps, and the rest using iteratively retrained versions of $C$.
Aside from the variety of objects, each robot differed slightly in terms of wear-and-tear, as well as the camera pose. 
The outcome of the grasp attempt was determined automatically.
The particular objects in front of each robot were regularly rotated to increase the diversity of the dataset. Some examples of grasping images from the camera's viewpoint are shown in Figure~\ref{fig:procedural_training}d.

When trained on the entire real dataset, the best CNN used in the approach outlined above achieved successful grasps $67.65\%$ of the time. Levine \etal~\cite{levine2016learning} reported an additional increase to $77.18\%$ from also including 2.7 million images from a different robot. We excluded this additional dataset for the sake of a more controlled comparison, so as to avoid additional confounding factors due to domain shift within the real-world data. Starting from the Kuka dataset, our experiments study the effect of adding simulated data and of reducing the number of real world data points by taking subsets of varying size (down to only $93,841$ real world images, which is $1\%$ of the original set).

\subsection{Domain Adaptation}
As part of our proposed approach we use two domain adaptation techniques: domain-adversarial training  and  pixel-level domain adaptation.
Ganin \etal~\cite{ganin2016domain} introduced domain--adversarial neural
networks (DANNs), an architecture trained to extract domain-invariant yet expressive features. DANNs were primarily tested in the unsupervised domain adaptation scenario, in the absence of any labeled target domain samples, although they also showed promising results in the semi-supervised regime~\cite{bousmalis2016domain}.
Their model's first few layers are shared by two modules: the first
predicts task-specific labels when provided with source data
while the second is a separate \textit{domain classifier} trained to predict the domain $\hat{d}$ of its inputs. 
The DANN loss is the cross-entropy loss for the domain prediction task:
${\mathcal {L}_\mathrm{DANN} = \;\sum_{i=0}^{N_s+N_t} \left\{d_i \log \hat{d}_i \;+\;  (1-d_i) \log(1- \hat{d}_i)\right \}}$,
where $d_i \in \{0,1\}$ is the ground truth domain label for sample $i$, and $N_s, N_t$ are the number of source and target samples.

The shared layers are trained to maximize $\mathcal {L}_\mathrm{DANN}$, while the domain classifier is trained adversarially to minimize it. 
This minimax optimization is implemented by a gradient reversal layer (GRL). 
The GRL has the same output as the identity function,
but negates the gradient during backprop.
This lets us compute the gradient for both the domain classifier and the shared feature extractor in a single backward pass.
The task loss of interest is simultaneously optimized with respect to the shared layers, which grounds the shared features to be relevant to the task.

While DANN makes the features extracted from both domains similar, the goal in \textit{pixel-level} domain adaptation ~\cite{bousmalis2017pixelda,shrivastava2017learning,zhu2017unpaired,taigman2017unsupervised} is to learn a generator function $G$ that maps images from a source to a target domain at the input level. This approach decouples the process of domain adaptation from the process of task-specific predictions, by adapting the images from the source domain to make them appear as if they were sampled from the target domain. Once the images are adapted, they can replace the source dataset and the relevant task model can be trained as if no domain adaptation were required.
 Although all these methods are similar in spirit, we use ideas primarily from PixelDA~\cite{bousmalis2017pixelda} and SimGAN~\cite{shrivastava2017learning}, as they are more suitable for our task. These models are particularly effective if the goal is to maintain the semantic map of original and adapted synthetic images, as the transformations are primarily low-level: the methods make the assumption that the differences between the domains are primarily low-level (due to noise, resolution, illumination, color) rather than high-level (types of objects, geometric variations, etc).

More formally, let ${\bs X}^s = \{{\bs x}_i^s, {\bs y}_i^s\}_{i=0}^{N^s}$
represent a dataset of $N^s$ samples from the source domain and let
${\bs X}^t = \{{\bs x}_i^t,  {\bs y}_i^t\}_{i=0}^{N^t}$ represent a
dataset of $N^t$ samples from the target domain.
The generator function
$G({\bs x}^s; {\bs \theta}_G) \rightarrow {\bs x^f}$,
parameterized by ${\bs \theta}_G$, maps a source image 
${\bs x}^s \in {\bs X}^s$ 
to an adapted, or fake, image ${\bs x^f}$. This function is learned with the help of an adversary, a discriminator function $D({\bs x};{\bs \theta}_D)$ that outputs the likelihood $d$ that a given image $\bs x$ is a real-world sample. Both $G$ and $D$ are trained using the standard adversarial objective~\cite{goodfellow2014generative}.
Given
the learned generator function $G$, it is possible to create a new dataset
${\bs X}^f = \left\{ G({\bs x}^s), {\bs y}^s\right\}$.
Finally, given an adapted dataset ${\bs X}^f$, the task-specific
model can be trained as if the training and test data were 
from the same distribution. 

PixelDA was evaluated in simulation-to-real-world transfer. However, the 3D models used by the renderer in~\cite{bousmalis2017pixelda} were very high-fidelity scans of the objects in the real-world dataset. In this work we examine for the first time how such a technique can be applied in situations where \textsl{(a)} no 3D models for the objects in the real-world are available and \textsl{(b)} the system is supposed to generalize to yet another set of previously unseen objects in the actual real-world grasping task.
Furthermore, we use images of $472\times472$, more than double the resolution in~\cite{bousmalis2017pixelda, shrivastava2017learning}. This makes learning the generative model $G$ a much harder task and requires significant changes compared to previous work: the architecture of both $G$ and $D$, the GAN training objective, and the losses that aid with training the generator (content-similarity and task losses) are different from the original implementations, resulting in a novel model evaluated under these new conditions. 

\section{Our Approach}
One of the aims of our work is to study how final grasping performance is affected by the 3D object models our simulated experience is based on, the scene appearance and dynamics in simulation, and the way simulated and real experience is integrated for maximal transfer. In this section we outline, for each of these three factors, our proposals for effective simulation-to-real-world transfer for our task.

\subsection{Grasping in Simulation}%
\label{sect:methods}
A major difficulty in constructing simulators for robotic learning is to ensure diversity sufficient for effective generalization to real-world settings.
In order to evaluate simulation-to-real world transfer, we used one dataset of real-world grasp attempts (see Sect.~\ref{sec:hand-eye}), and multiple such datasets in simulation. 
For the latter, we built a basic virtual environment based on the Bullet physics simulator and the simple renderer that is shipped with it~\cite{coumans2017}. The environment emulates the Kuka hardware setup by simulating the physics of grasping and by rendering what a camera mounted looking over the Kuka shoulder would perceive: the arm, the bin that contains the object, and the objects to grasp in scenes similar to the ones the robot encounters in the real world. 

\begin{figure}[t]
\centering
\begin{subfigure}[b]{.32\linewidth}
\includegraphics[width=\textwidth]{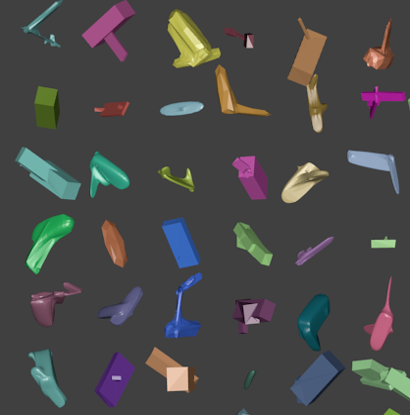}
\caption{Procedural}
\end{subfigure}
\begin{subfigure}[b]{.32\linewidth}
\includegraphics[width=\linewidth]{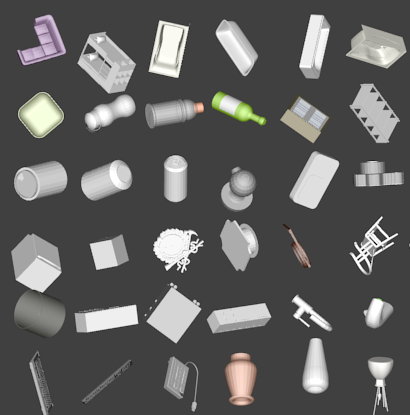}
\caption{ShapeNet~\cite{shapenet}}
\end{subfigure}
\begin{subfigure}[b]{.32\linewidth}
\includegraphics[width=\linewidth]{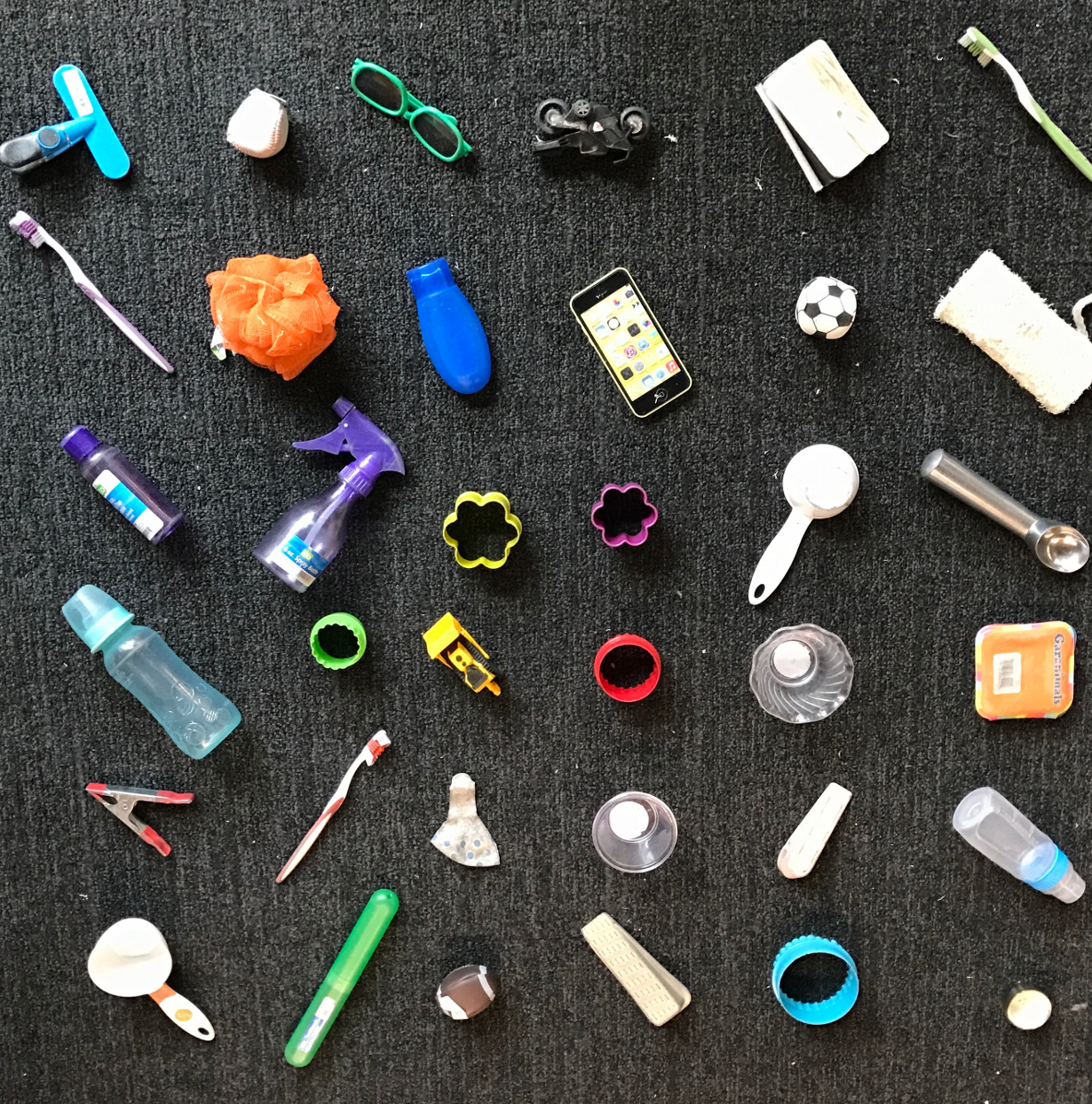}
\caption{Real}
\end{subfigure}
\caption{Comparison of \textsl{(a)} some of our $1,000$ procedural, \textsl{(b)} some of the $51,300$ ShapeNet objects, both used for data collection in simulation, and the \textsl{(c)} $36$ objects we used only for evaluating grasping in the real-world, that were not seen during training. The variety of shapes, sizes, and material properties makes the test set very challenging. \vspace{-.2in}}
\label{fig:sim_objects}
\end{figure}

A central question here is regarding the realism of the 3D models used for the objects to grasp.
To answer it, we evaluate two different sources of objects in our experiments: \textsl{(a)} procedurally generated random geometric shapes and \textsl{(b)} realistic objects obtained from the publicly-available ShapeNet~\cite{shapenet} 3D model repository. 
We procedurally generated $1,000$ objects by attaching rectangular prisms at random locations and orientations, as seen in Fig.~\ref{fig:sim_objects}a.
We then converted the set of prisms to a mesh using an off-the-shelf renderer, Blender, and applied a random level of smoothing.
Each object was given UV texture coordinates
and random colors.
For our Shapenet-based datasets, we used the ShapeNetCore.v2~\cite{shapenet} collection of realistic object models, shown in Figure~\ref{fig:sim_objects}b.
This particular collection contains $51,300$ models in 55 categories of household objects, furniture, and vehicles.
We rescaled each object to a random graspable size with a maximum extent between 12cm and 23cm (real-world objects ranged from 4cm to 20cm in length along the longest axis) and gave it a random mass between 10g and 500g, based on the approximate volume of the object.

Once the models were imported into our simulator, we collected our simulation datasets via a similar process to the one in the real world, with a few differences. As mentioned above, the real-world dataset was collected by using progressively better grasp prediction networks. These networks were swapped for better versions manually and rather infrequently~\cite{levine2016learning}.
In contrast to the 6 physical Kuka IIWA robots that were used to collect data in the real world, we used 1,000 to 2,000 simulated arms at any given time to collect our synthetic data, and the models that were used to collect the datasets were being updated continuously by an automated process. This resulted in datasets that were collected by grasp prediction networks of varying performance, which added diversity to the collected samples. 
After training our grasping approach in our simulated environment, the simulated robots were successful on 70\%-90\% of the simulated grasp attempts.
Note that all of the grasp success prediction models used in our experiments were trained from scratch using these simulated grasp datasets.

\subsection{Virtual Scene Randomization}
\label{sect:scene-randomization}
Another important question is whether randomizing the visual appearance and dynamics in the scene affects grasp performance and in what way. One of the first kind of diversities we considered was the addition of  $\epsilon$~cm, where $\epsilon \sim {\cal N}(0, 1)$,  to the horizontal components of the motor command. This improved real grasp success in early experiments, so we added this kind of randomization for all simulated samples.
Adding this noise to real data did not help. To further study the effects of virtual scene randomization, we built datasets with four different kinds of scene randomization: \textsl{(a) No randomization}: Similar to real-world data collection, we only varied camera pose, bin location, and used 6 different real-world images as backgrounds; 
\textsl{(b) Visual Randomization}: We varied tray texture, object texture and color, robot arm color, lighting direction and brightness; \textsl{(c) Dynamics Randomization}: We varied object mass, and object lateral/rolling/spinning friction coefficients; and \textsl{(d) All}: both visual and dynamics randomization.

\subsection{Domain Adaptation for Vision-Based Grasping} 
\label{sec:pixelda}
As mentioned in Sect.~\ref{sect:related}, there are two primary types of methods used for domain adaptation: feature-level, and pixel-level. 
Here we propose a feature-level adaptation method and a novel pixel-level one, which we call GraspGAN. Given original synthetic images, GraspGAN produces adapted images that look more realistic. We subsequently use the trained generator from GraspGAN as a fixed module that adapts our synthetic visual input, while performing feature-level domain adaptation on extracted features that account for both the transferred images and synthetic motor command input.

\begin{figure*}[ht]
\centering
\includegraphics[width=\linewidth]{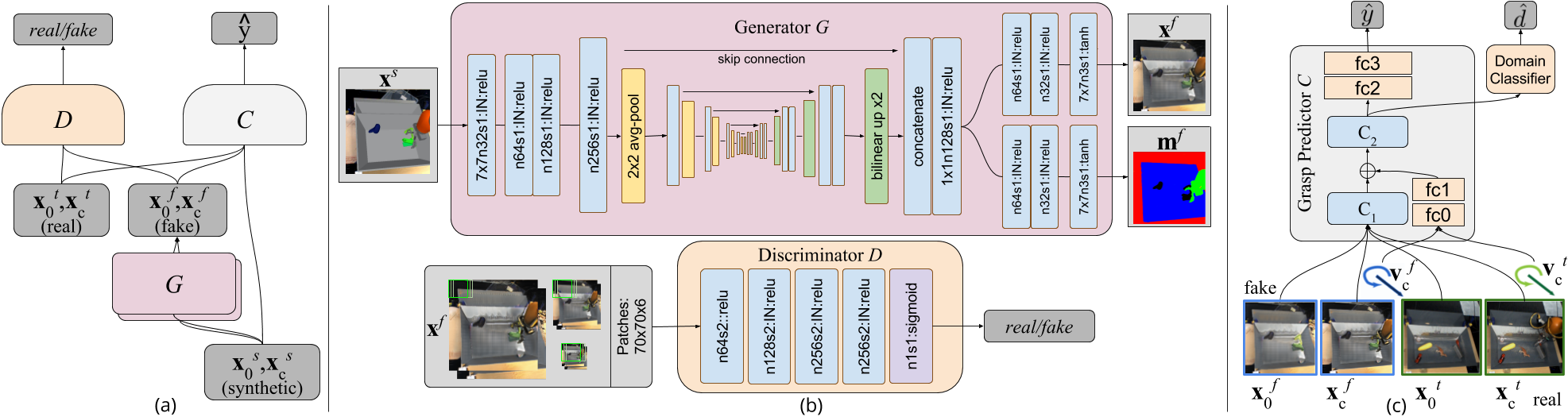}
\caption{{\bf Our proposed approach:}  {\bf (a)} Overview of our pixel-level domain adaptation model, {GraspGAN}. Tuples of images from the simulation ${\bs x}^s$ are fed into the generator $G$ to produce realistic versions ${\bs x}^f$. The discriminator $D$ gets unlabeled real world images ${\bs x}^t$ and ${\bs x}^f$ and is trained to distinguish them. Real and adapted images are also fed into the grasp success prediction network $C$, trained in parallel (motion commands $\bs v$ are not shown to reduce clutter). $G$, thus, gets feedback from $D$ and $C$ to make adapted images look real and maintain the semantic information. {\bf (b)} Architectures for $G$ and $D$. Blue boxes denote convolution/normalization/activation-layers, where $n64s2{:}IN{:}relu$ means $64$ filters, stride $2$, instance normalization $IN$ and $relu$ activation. Unless specified all convolutions are $3\times3$ in $G$ and $4\times4$ in $D$. {\bf (c)} DANN model: $C_1$ has 7 conv layers and $C_2$ has 9 conv layers. Further details can be found in~\cite{levine2016learning}. Domain classifier uses GRL and two 100 unit layers. \vspace{-5mm}}
\label{fig:graspgan}
\end{figure*}

For our feature-level adaptation technique we use a DANN loss on the last convolutional layer of our grasp success prediction model $C$, as shown in Fig.~\ref{fig:graspgan}c.
In preliminary experiments we found that using the DANN loss on this layer yielded superior performance compared to applying it at the activations of other layers. We used the domain classifier proposed in~\cite{ganin2016domain}. One of the early research questions we faced was what the interaction of batch normalization (BN) ~\cite{ioffe2015batch} with the DANN loss would be, as this has not been examined in previous work. We use BN in every layer of $C$ and in a na\"{i}ve implementation of training models with data from two domains, a setting we call {\bf na\"{i}ve mixing}, batch statistics are calculated without taking the domain labels of each sample into account. 
However,
the two domains are bound to have different statistics, which
means that calculating and using them separately for
simulated and real-world data while using the same
parameters for $C$ might be beneficial.
We call this way of training data from two domains {\bf domain-specific batch normalization (DBN) mixing}, and show it is a useful tool for domain adaptation, even when a DANN loss is not used.

In \textbf{our pixel-level domain adaptation model, GraspGAN}, shown in Fig.~\ref{fig:graspgan}, $G$ is a convolutional neural network that follows a U-Net 
architecture~\cite{ronneberger2015unet}, and uses average pooling for downsampling, bilinear upsampling, concatenation and $1\times1$ convolutions for the U-Net skip connections, and instance normalization~\cite{ulyanov2016instance}.  Our discriminator $D$ is a patch-based~\cite{isola2016image} CNN with 5 convolutional layers, with an effective input size of $70\times70$. It is fully convolutional on 3 scales 
($472\times472$, $236\times236$, and $118\times118$) 
of the two input images, ${\bs x}^s_0$ and ${\bs x}^s_c$,  stacked into a 6 channel input, producing domain estimates for all patches which are then combined to compute the joint discriminator loss.
This {\bf novel multi-scale patch-based discriminator} design can learn to assess both global consistency of the generated image, as well as realism of local textures. 
Stacking the channels of the two input images enables the discriminator to recognize relationships between the two images, so it can encourage the generator to respect them (\eg, paint the tray with the same texture in both images, but insert realistic shadows for the arm).
Our task model $C$ is the grasp success prediction CNN from~\cite{levine2016learning}.

To train  GraspGAN, we employ a least-squares generative adversarial objective (LSGAN)~\cite{mao2016least} 
to encourage $G$ to produce realistic images. During training, our generator
$G({\bs x}^{s}; {\bs \theta}_G) \rightarrow {\bs x^f}$
maps synthetic images ${\bs x}^s$ to adapted images ${\bs x^f}$, by individually passing ${\bs x}^s_0$ and ${\bs x}^s_c$ through two instances of the generator network displayed in Figure~\ref{fig:graspgan}. 
Similar to traditional GAN training, we perform optimization in alternating steps by minimizing the following loss terms \wrt the parameters of each sub-network:
\begin{align}
\min_{{\bs \theta}_G}&\lambda_g{\cal L}_{gen}(G, D)+\lambda_{tg}{\cal L}_{task}(G, C)+\lambda_c{\cal L}_{content}(G) \; \\
\min_{{\bs \theta}_D,{\bs \theta}_C}&\lambda_d{\cal L}_{discr}(G, D)+\lambda_{td}{\cal L}_{task}(G, C),
\end{align}
where ${\cal L}_{gen}$ and ${\cal L}_{discr}$ are the LSGAN generator and discriminator losses, ${\cal L}_{task}$ is the task loss,  ${\cal L}_{content}$ is the content-similarity loss, and $\lambda_g$, $\lambda_d$, $\lambda_{tg}$, $\lambda_{td}$, $\lambda_c$, the respective weights.
The LSGAN discriminator loss is the $L2$ distance between its likelihood output $\hat{d}$ and the domain labels $d=0$ for fake and $d=1$ for real images, while for the generator loss the label is flipped, such that there is a high loss if the disciminator predicts $\hat{d}=0$ for a generated image.
The task loss measures how well the network $C$ predicts grasp success on transferred and real examples by calculating the binomial cross-entropy of the labels $y_i$.

It is of utmost importance that the GraspGAN generator, while making the input image look like an image from the real world scenario, does not change the semantics of the simulated input, for instance by drawing the robot's arm or the objects in different positions. Otherwise, the information we extract from the simulation in order to train the task network would not correspond anymore to the generated image. We thus devise several additional loss terms, accumulated in ${\cal L}_{content}$, to help anchor the generated image to the simulated one on a semantic level.
The most straightforward restriction is to not allow the generated image to deviate much from the input. To that effect we use the PMSE loss, also used by~\cite{bousmalis2017pixelda}.
We also leverage the fact that we can have semantic information about every pixel in the synthetic images by computing segmentation masks ${\bs m}^f$ of the  corresponding rendered images for the background, the tray, robot arm, and the objects.
We use these masks by training our generator $G$ to also produce ${\bf m}^f$ as an additional output for each adapted image, with a standard $L2$ reconstruction loss. Intuitively, it forces the generator to extract semantic information about all the objects in the scene and encode them in the intermediate latent representations. This information is then available during the generation of the output image as well.
Finally, we additionally implement a loss term that provides more dense feedback from the task tower than just the single bit of information about grasp success. We encourage the generated image to provide the same semantic information to the task network as the corresponding simulated one by penalizing differences in activations of the final convolutional layer of $C$ for the two images.
This is similar in principle to the perceptual loss~\cite{johnson2016perceptual} that uses the activations of an ImageNet-pretrained VGG model as a way to anchor the restylization of an input image. In contrast, here $C$ is trained at the same time, the loss is specific to our goal, and it helps preserve the semantics in ways that are relevant to our prediction task.

\section{Evaluation}
This section aims to answer the following research questions: \textsl{(a)} is the use of simulated data from a low quality simulator aiding in improving grasping performance in the real world? \textsl{(b)} is the improvement consistent with varying amounts of real-world labeled samples? \textsl{(c)} how realistic do graspable objects in simulation need to be? \textsl{(d)} does randomizing the virtual environment affect simulation-to-real world transfer, and what are the randomization attributes that help most? \textsl{(e)} does domain adaptation allow for better utilization of simulated grasping experience?
\begin{figure}[b]
\centering
\includegraphics[width=\linewidth]{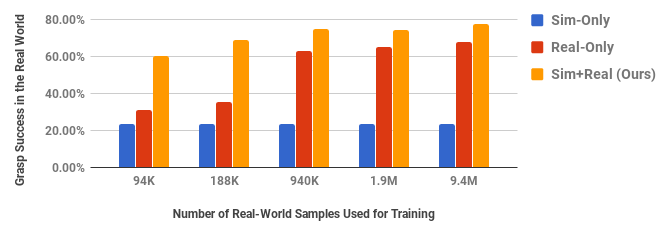}
\caption{The effect of using 8 million simulated samples of procedural objects with no randomization and various amounts of real data, for the best technique in each class.\vspace{-5mm}}
\label{fig:results_teaser}
\end{figure}

In order to answer these questions, we evaluated
a number of different ways for training a grasp success prediction model $C$ with simulated data and domain adaptation\footnote{Visit \url{https://goo.gl/G1HSws} for our supplementary video.}. When simulated data was used, the number of simulated samples was always approximately 8 million.
We follow the grasp success evaluation protocol described by Levine \etal~\cite{levine2016learning}.  
We used 6 Kuka IIWA robots for our real-world experiments and a test set consisting of the objects shown in Fig.~\ref{fig:sim_objects}c,
the same used in~\cite{levine2016learning}, with 6 different objects in each bin for each robot.
These objects were not included in the real-world training set and were not used in any way when creating our simulation datasets.
Each robot executes 102 grasps, for a total of 612 test grasps for each evaluation.
During execution, each robot picks up objects from one side of the
bin and drops them on the other, alternating every 3 grasps. This prevents the model from repeatedly grasping the same object.
Optimal models were selected by using the accuracy of $C$ on a held-out validation set of $94,000$ real samples.

The first conclusion from our results is that simulated data from an off-the-shelf simulator always aids in improving vision-based real-world grasping performance. As one can see in Fig.~\ref{fig:results_teaser}, which shows the real grasp success gains by incorporating simulated data from our procedurally-generated objects, using our simulated data significantly and consistently improves real-world performance regardless of the number of real-world samples. 

We also observed that we do not need realistic 3D models to obtain these gains. We compared the effect of using random, procedurally-generated shapes and ShapeNet objects in combination with 10\% of the real-world data, under all randomization scenarios.
As shown in Table~\ref{tab:randomization-results} we found that using procedural objects is the better choice in all cases. This finding has interesting implications for simulation to real-world transfer, since content creation is often a major bottleneck in producing generalizable simulated data.
Based on these results, we decided to use solely procedural objects for the rest of our experiments.

\begin{table}[t]
\centering
\caption{{\bf The effect of our choices for simulated objects and randomization in terms of grasp success.} We compared the performance of models trained jointly on grasps of procedural vs ShapeNet objects with 10\% of the real data. Models were trained with DANN and DBN mixing.
}
\begin{tabular}{| l | c | c | c | c |}
\hline
{\bf Randomization} & None & Visual & Dynamics & Both \\\hline\hline
{\bf Procedural} & {\bf 71.93\%} & {\bf 74.88\%} & {\bf 73.95\%} & {\bf 72.86\%} \\ \hline
{\bf ShapeNet} & 69.61\% & 68.79\% & 68.62\% & 69.84\%  \\ \hline
\end{tabular}
\label{tab:randomization-results} \vspace{-0.2in}
\end{table}

Table~\ref{tab:grasp-results} shows our main results: the grasp success performance for different combinations of simulated data generation and domain adaptation methods, and with different quantities of real-world samples. The different settings are:  \textbf{\textit{Real-Only}}, in which the model is given only real data; \textit{Na\"{i}ve Mixing (\textbf{Na\"{i}ve Mix})}: Simulated samples generated with no virtual scene randomization are mixed with real-world samples such that half of each batch consists of simulated images; \textit{DBN Mixing \& Randomization (\textbf{Rand.})}: The simulated dataset is generated with visual-only randomization. The simulated samples are mixed with real-world samples as in the naive mixing case, and the models use DBN; \textit{DBN Mixing \& DANN (\textbf{DANN})}: Simulated samples are generated with no virtual scene randomization and the model is trained with a domain-adversarial method with DBN; \textit{DBN Mixing, DANN \& Randomization (\textbf{DANN-R})}: Simulated samples are generated with visual  randomization and the model is trained with a domain-adversarial method with DBN; \textit{GraspGAN, DBN Mixing \& DANN (\textbf{GraspGAN})}: The non-randomized simulated data is first refined with a GraspGAN generator, and the refined data is used to train a DANN with DBN mixing. The generator is trained with the same real dataset size used to train the DANN. See Figure~\ref{fig:intro_fig}b for examples.

Table~\ref{tab:grasp-results} shows that using visual randomization with DBN mixing improved upon na\"{i}ve mixing with no randomization experiments across the board.
The effect of visual, dynamics, and combined randomization for both procedural and ShapeNet objects was evaluated by using $10\%$ of the real data available. Table~\ref{tab:randomization-results} shows that using only visual randomization slightly improved grasp performance for procedural objects, but the differences were generally not conclusive.

\begin{table}[t]
\centering
\caption{Real grasp performance when no labeled real examples are available. Method names explained in the text.}
\begin{tabular}{| c | c | c |}
\hline
{\bf Sim-Only} & {\bf Rand.} &  {\bf GraspGAN} \\ \hline  \hline
23.53\% & 35.95\% &  {\bf 63.40\% }\\ \hline
\end{tabular}
\label{tab:no-label-results} \vspace{-3mm}
\end{table}

In terms of domain adaptation techniques, our proposed hybrid approach of combining our GraspGAN and DANN performs the best in most cases, and shows the most gains in the lower real-data regimes. Using DANNs with DBN Mixing performed better than na\"{i}ve mixing in most cases. However the effect of DANNs on Randomized data was not conclusive, as the equivalent models produced worse results in 3 out of 5 cases. We believe the most interesting results however, are the ones from our experiments with no labeled real data. 
We compared the best domain adaptation method (GraspGAN), against a model trained on simulated data with and without randomization.
We trained a GraspGAN on all 9 million real samples, without using their labels. Our grasping model was then trained only on data refined by $G$. Results in Table~\ref{tab:no-label-results} show that the unsupervised adaptation model
outperformed not only sim-only models with and without randomization but also a real-only model with 939,777 labeled real samples.

\begin{table}[t]
\centering
\caption{Success of grasping $36$ diverse and unseen physical objects of all our methods trained on different amounts of real-world samples and 8 million simulated samples with procedural objects. Method names are explained in the text.
}
\begin{tabular}{| l | c | c | c | c | c | c}
\hline
\multirow{2}{*}{\bf Method} & \bf All &\bf 20\% & \bf 10\%  & \bf 2\%  & \bf 1\% \\ 
&\bf 9,402,875&\bf 1,880,363& \bf 939,777 & \bf 188,094  & \bf 93,841 \\ 
\hline \hline
Real-Only       & 67.65\% & 64.93\% & 62.75\% & 35.46\% & 31.13\%\\ \hline\hline
Na\"{i}ve Mix.    & 73.63\% & 69.61\% & 65.20\% & 58.38\% & 39.86\% \\ \hline
Rand.   & 75.58\%     & 70.16\%     & 73.31\% & 63.61\% & 50.99\% \\ \hline
DANN           & 76.26\% & 68.12\% & 71.93\% & 61.93\% & 59.27\%\\ \hline
DANN-R.    & 72.60\% & 66.46\% & {\bf 74.88\%} & 63.73\%& 43.81\% \\ \hline
GraspGAN & {\bf 76.67}\%     & {\bf 74.07\%}     & 70.70\% & {\bf 68.51\%}& {\bf 59.95\%} \\ \hline
\end{tabular}
\label{tab:grasp-results} \vspace{-5mm}
\end{table}

Although our absolute grasp success numbers are consistent with the ones reported in~\cite{levine2016learning}, some previous grasping work reports higher absolute grasp success. 
However, we note the following: \textsl{(a)} our goal in this work is not to show that we can train the best possible grasping system, but that for the same amount of real-world data, the inclusion of synthetic data can be helpful; we have relied on previous work~\cite{levine2016learning} for the grasping approach used; \textsl{(b)} our evaluation was conducted on a diverse and challenging range of objects, including transparent bottles, small round objects, deformable objects, and clutter; and \textsl{(c)} the method uses only monocular RGB images from an over-the-shoulder viewpoint, without depth or wrist-mounted cameras. These make our setup considerably harder than most standard ones.

\section{Conclusion}
In this paper, we examined how simulated data can be incorporated into a learning-based grasping system to improve performance and reduce data requirements. We study grasping from over-the-shoulder monocular RGB images, a particularly challenging setting where depth information and analytic 3D models are not available. This presents a challenging setting for simulation-to-real-world transfer, since simulated RGB images typically differ much more from real ones compared to simulated depth images. We examine the effects of the nature of the objects in simulation, of randomization, and of domain adaptation. We also introduce a novel extension of pixel-level domain adaptation that makes it suitable for use with high-resolution images used in our grasping system. Our results indicate that including simulated data can drastically improve the vision-based grasping system we use, achieving comparable or better performance with 50 times fewer real-world samples. Our results also suggest that it is not as important to use realistic 3D models for simulated training. Finally, our experiments indicate that our method can provide plausible transformations of synthetic images, and that including domain adaptation substantially improves performance in most cases.

Although our work demonstrates very large improvements in the grasp success rate when training on smaller amounts of real world data, there are a number of limitations. 
Both of the adaptation methods we consider focus on invariance, either transforming simulated images to look like real images, or regularizing features to be invariant across domains. These features incorporate both appearance and action, due to the structure of our network, but no explicit reasoning about physical discrepancies between the simulation and the real world is done.
We did consider randomization of dynamics properties, and show it is indeed important.
Several recent works have looked at adapting to physical discrepancies explicitly~\cite{christiano2016transfer,rajeswaran2016epopt,yu2017preparing},
and incorporating these ideas into grasping is an exciting avenue for future work.
Our approach for simulation to real world transfer only considers monocular RGB images, though extending this method to stereo and depth images would be straightforward.
Finally, the success rate reported in our experiments still has room for improvement, and we expect further research in this area will lead to even better results.
The key insight from our work comes from the comparison of the different methods: we are not aiming to propose a novel grasping system, but rather to study how incorporating simulated data can improve an existing one.
\vspace{-1mm}

%
\section*{Acknowledgments}  

The authors thank John-Michael Burke for overseeing the robot operations. The authors also thank Erwin Coumans, Ethan Holly, Dmitry Kalashnikov, Deirdre Quillen, and Ian Wilkes for contributions to the development of our grasping system and supporting infrastructure.

{\small
\bibliographystyle{bibtex/IEEEtran}
\bibliography{bibtex/sim2real}
}
\end{document}